\documentclass[
]{ceurart}

\usepackage{listings}
\usepackage{color,soul}
\begin{document}

\copyrightyear{2022}
\copyrightclause{Copyright for this paper by its authors.
  Use permitted under Creative Commons License Attribution 4.0
  International (CC BY 4.0).}

\conference{De-Factify 4.0 : Fourth Workshop on Multimodal Fact Checking and Hate Speech Detection
February, 2025, co-located with AAAI 2025 | Philadelphia, Pennsylvania, USA}

\title{Sarang at DEFACTIFY 4.0: Detecting AI-Generated Text Using Noised Data and an Ensemble of DeBERTa Models}



\author[1]{Avinash Trivedi}[
email=avinashtrivedi.2008@gmail.com
]
\cormark[1]
\address[1]{National Institute of Technology, Tiruchirapalli,India}
\author[1]{Sangeetha Sivanesan}[]
\cortext[1]{Corresponding author}

\begin{abstract}
This paper presents an effective approach to detect AI-generated text, developed for the Defactify 4.0 shared task at the fourth workshop on multimodal fact checking and hate speech detection. The task consists of two subtasks: Task-A, classifying whether a text is AI generated or human written, and Task-B, classifying the specific large language model that generated the text.  Our team (\textit{Sarang}) achieved the 1st place in both tasks with F1 scores of 1.0 and 0.9531, respectively. The methodology involves adding noise to the dataset to improve model robustness and generalization. We used an ensemble of DeBERTa models to effectively capture complex patterns in the text. The result indicates the effectiveness of our noise-driven and ensemble-based approach, setting a new standard in AI-generated text detection and providing guidance for future developments.
\end{abstract}

\begin{keywords}
  AI-generated text detection \sep
  DeBERTa \sep
  Noise \sep
  De-Factify 4.0@AAAI2025
\end{keywords}

\maketitle

\section{Introduction}

Large Language Models (LLMs), such as ChatGPT \cite{schulman2022chatgpt}, are really good at writing long pieces of text that sound very human. While these developments have various beneficial applications, they also raise concerns about potential misuse, such as the automatic creation of fake news articles and academic contents \cite{natureChatGPTWrites}. To address these risks, various algorithms have been developed to detect AI-generated text, which include watermarking techniques \cite{kirchenbauer2024watermarklargelanguagemodels}, tools like GPTZero \cite{gptzeroDetectorOriginal}, DetectGPT \cite{mitchell2023detectgptzeroshotmachinegeneratedtext}, and OpenAI's text classifier \cite{openai-text}. 

The task of detecting AI-generated text is inherently challenging, as recent research \cite{sadasivan2024aigeneratedtextreliablydetected} highlights the increasing sophistication of newer, more capable LLMs. Early studies demonstrated that humans struggle to tell if something was written by a computer or a human. Given the ethical implications and the complexity of the problem, creating robust detection systems remains an active area of research. 
The Defactify 4.0 shared task \footnote{\url{https://defactify.com/ai_gen_txt_detection.html}} \cite{roy-2025-defactify-overview-text}, part of the fourth workshop on multimodal fact-checking and hate speech detection, featured two subtasks: Task-A focused on distinguishing between AI-generated and human-authored text, while Task-B aimed to identify the specific LLM responsible for generating the text. This paper proposes an ensemble based DeBERTa model, trained and validated on noisy dataset to make the model more robust. This work highlights how adding noise to the dataset makes the model remain resilient to disturbances. It captures features invariant under perturbations and demonstrates significantly improved robustness against such disturbances.

The rest of the paper is as follows. Section 2 contains related work, section 3 describes the dataset, section 4 describes our methodology, section 5 contains experimental results and section 6 includes conclusions and future work.

\section{Related work}
Recent advancements have demonstrated significant progress in methods for detecting AI-generated text. These methods broadly fall into three categories: statistical approaches, classifier-based detectors, and watermarking techniques.

\textbf{Traditional statistical detection methods} leverage metrics such as entropy, perplexity, and n-gram frequency to identify differences in linguistic patterns between human and machine-generated text \cite{Lavergne2008DetectingFC,Gehrmann2019GLTRSD}. A recent innovation, DetectGPT \cite{mitchell2023detectgptzeroshotmachinegeneratedtext}, builds on these principles, focusing on the negative curvature areas of a model's log probability. By generating and comparing perturbed variations of text, DetectGPT determines its likelihood of being machine-generated based on log probabilities. This method achieves significantly higher AUROC scores compared to other zero-shot detection approaches, making it a notable advancement in statistical detection techniques.

\textbf{Classifier-based detection methods} are commonly employed in identifying fake news and misinformation \cite{schildhauer2022fake,zou2021ai}. OpenAI, for example, fine-tuned a GPT model using datasets from Wikipedia, WebText, and human-labeled samples to create a classifier capable of discerning machine-generated text. This model combines automated classification with human evaluation, demonstrating its efficacy in detecting AI-generated content. Such advancements contribute to mitigating the spread of misinformation and improving societal trust in online content \cite{kshetri2022deep}.

\textbf{Watermark-Based Identification} has emerged as a compelling alternative for machine-generated text detection. Historically used in image processing for copyright protection and data hiding \cite{langelaar2000watermarking,wmark_old1}, watermarking techniques have recently been adapted for natural language. \cite{kirchenbauer2023watermark} proposed a novel watermarking approach that utilizes language model logits to embed invisible watermarks in text. This method categorizes tokens into \textit{green} and \textit{red} lists, guiding token selection to create patterns that are imperceptible to human readers. These advancements not only enhance content authentication and copyright enforcement but also pave the way for secure communication, digital rights management, and privacy protection.

While existing methods effectively identify unaltered LLM-generated content, their reliability against user-modified versions remains underexplored. Research shows that even small changes can significantly weaken the performance of these detection techniques. We proposed a noise-driven, DeBERTa based ensemble approach to address the issue, as it remains largely unaffected by disturbances and highlight differences between human and LLM-generated text. This method improves robustness in detecting perturbed LLM-generated content.

Our method is inspired from \cite{karpukhin2019training,wei-zou-2019-eda,xie2017data}, where \cite{karpukhin2019training} observed that training machine translation models on a balanced mix of simple synthetic noise enhances robustness to character-level variations, such as typos, without compromising performance on clean text. Authors in \cite{wei-zou-2019-eda} introduces Easy Data Augmentation (EDA) four simple yet effective techniques, synonym replacement, random insertion, random swap, and random deletion. It significantly improves text classification performance, especially for smaller datasets, achieving comparable results with reduced training data.
Authors in \cite{xie2017data} highlights that adding noise can be beneficial for model generalization. It proposes two Noising technique, First is Unigram Noising, Which randomly replaces tokens in a sequence with words sampled from the unigram frequency distribution at a probability \( \gamma \), introducing corpus-wide diversity. Second is Blank Noising, Which replaces tokens with a placeholder token (“\_”) at a probability \( \gamma \), simulating missing context to enhance generalization. We used DeBERTa \cite{he2020deberta} as our base model, which is becoming popular in NLP because they can predict words using both the left and right context and are trained on a large amount of plain text from the internet.

\section{Dataset}
The dataset \cite{roy-2025-defactify-dataset-text} provided in this shared task consists of three columns namely Text, Label\_A and Label\_B, where Text is the AI or Human generated text, Label\_A denoting class 0 or 1 (Human/AI), Label\_B denotes one of the specific LLMs (Human\_story, gemma-2-9b, mistral-7B, qwen-2-72B, llama-8B, yi-large or GPT\_4-o) that generated the text. Table \ref{tab:data_distribution} contains dataset distribution. 

\begin{table}[h!]
\centering
\begin{tabular}{lcc}
\toprule
\textbf{Class} & \textbf{Validation Set Count} & \textbf{Training Set Count} \\ 
\midrule
Mistral-7B     & 1569                          & 7321                        \\ 
Llama-8B       & 1569                          & 7321                        \\ 
GPT\_4-o       & 1569                          & 7321                        \\ 
Qwen-2-72B     & 1569                          & 7321                        \\ 
Yi-Large       & 1569                          & 7321                        \\ 
Gemma-2-9B     & 1569                          & 7321                        \\ 
Human\_Story   & 1569                          & 7321                        \\ 
\midrule
\textbf{Total}       & \textbf{10983}                & \textbf{51247}        \\ 
\bottomrule
\end{tabular}
\caption{Data distribution for validation and training set}
\label{tab:data_distribution}
\end{table}

\section{Methodology}
\subsection{Finetuning of DeBERTa on Original Dataset}
We started by fine-tuning a set of DeBERTa models on original train set and validated on original val set to create a trustworthy baseline for our investigations. Table \ref{tab:baseline_score} provides a summary of the findings from these experiments. Out of all the evaluated configurations, the \textit{DeBERTa-v3-small} model performed the best on Task-A, showing the most promising result on test set. This suggests that the model is capable of successfully capturing the subtleties and patterns required to meet Task-A's requirements. 

While working on Task-B, we found that all models, including \textit{DeBERTa-v3-small}, exhibited signs of overfitting in spite of our various attempts. The training and testing performances diverged significantly as a result of this overfitting, which eventually resulted in less than ideal generalisation for Task-B. These results imply that in order to enhance performance on Task-B while preserving strong outcomes for Task-A, further tactics, such as regularisation schemes, data augmentation, or different modelling approaches, might be required.

\begin{table}[h!]
\centering
\begin{tabular}{llcc}
\toprule
\textbf{Data} & \textbf{Model} & \multicolumn{2}{c}{\textbf{Test F1-Score}} \\ 
\cmidrule(lr){3-4}
& & \textbf{Label-A} & \textbf{Label-B} \\ 
\midrule
& deberta-v3-xsmall & 0.9521 & 0.2418 \\ 
Original train/val & deberta-v3-small & \textbf{0.9985} & 0.2885 \\ 
& deberta-v3-base & 0.9515 & 0.4322 \\ 
\bottomrule
\end{tabular}
\caption{Baseline scores on test set}
\label{tab:baseline_score}
\end{table}

\subsection{Best performing system}

\begin{figure}
	\centering
		\includegraphics[scale=.15]{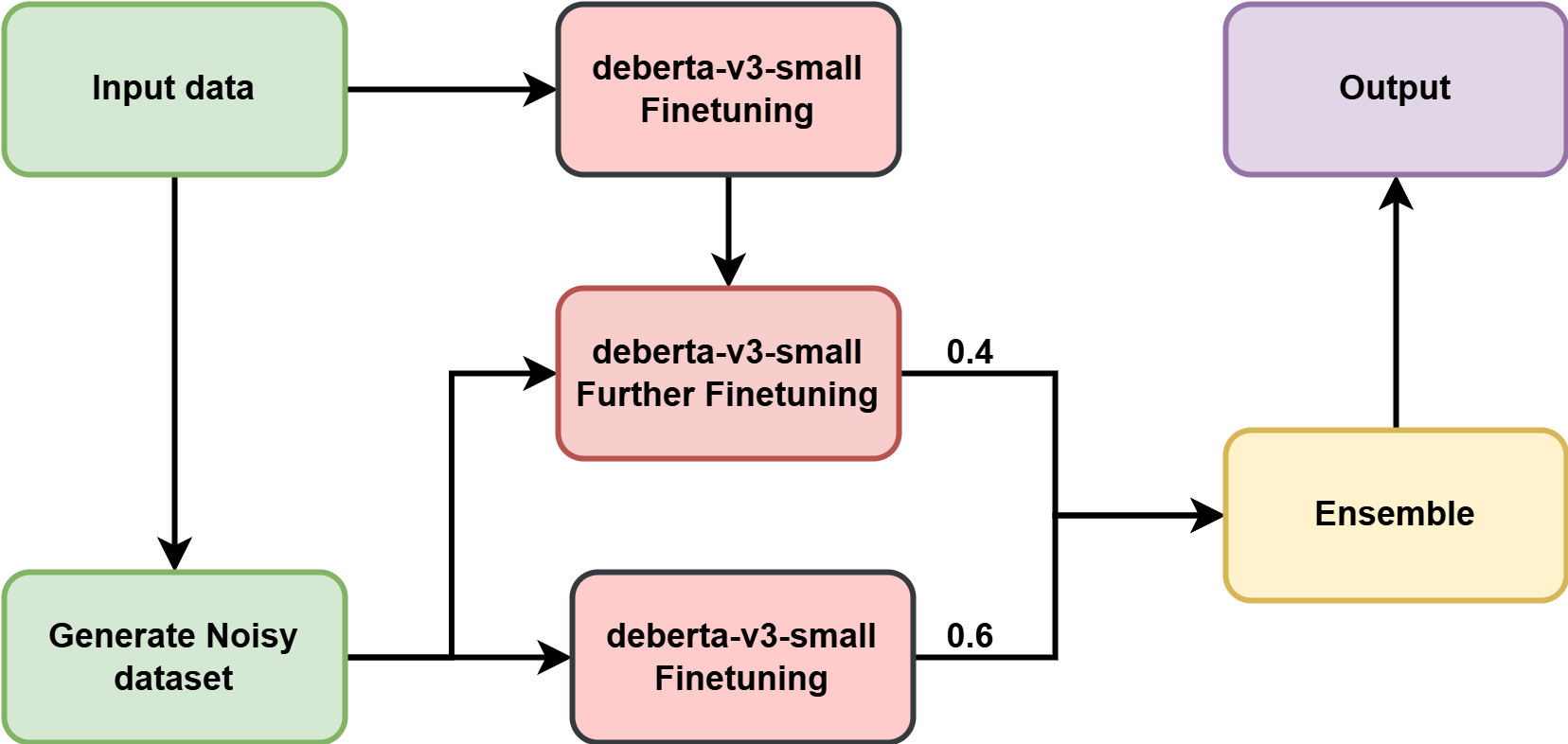}
	\caption{Ensemble based model architecture}
	\label{FIG:arch}
\end{figure}

Inspired from \cite{karpukhin2019training,wei-zou-2019-eda,xie2017data}, we implemented a data noising strategy to enhance the robustness of our language model. This technique, inspired by the authors' insights on the benefits of noise injection for smoothing, involves introducing controlled disruptions to the dataset. The architecture of our best-performing system is
illustrated in Fig \ref{FIG:arch}. We noised our dataset by injecting 10\% junk or garbled words into each data points as shown in Table \ref{tab:noised_data}. These junk words were randomly generated with lengths varying between 3 to 8 characters, ensuring the injected noise was both unpredictable and diverse. This approach mimics real-world scenarios where noisy or corrupted data is often encountered, enabling the model to learn more resilient representations.

Once the noisy dataset was prepared, we utilized it to fine-tune the DeBERTa model, a state-of-the-art transformer-based language model. By exposing the model to this noised data during fine-tuning, we aimed to evaluate its ability to generalize effectively and maintain performance in the presence of noisy inputs. We have finetuned \textit{deberta-v3-small} on original and noisy train data, also further finetune \textit{deberta-v3-small} on noisy train data which was earlier finetuned on original train dataset. We created a weighted ensemble (60:40) combining two models: one fine-tuned on noisy training data and the other sequentially fine-tuned on both original and noisy training data. Our primary goal was to test whether data noising could serve as a regularization technique to reduce overfitting and enhance the model's robustness.


\begin{table}[h!]
\centering
\begin{tabular}{p{7 cm}|p{7 cm}}
\hline
\textbf{Original Text} & \textbf{Noised text} \\ \hline
Photos of the Day: Greece and Elsewhere Migrants started small fires to keep warm near Idomeni, Greece, where they waited to cross the border into Macedonia. Photos of the Day: Greece and Elsewhere Greece and More — Pictures of the Day Slideshow controls                            & Photos of the Day: Greece and \hl{gkkas} Elsewhere Migrants \hl{nvvwe} started small fires to keep warm near Idomeni, Greece, where they waited to cross the border into Macedonia. Photos of the Day: \hl{visrqmy} Greece and Elsewhere Greece and More — Pictures of the Day Slideshow controls                     \\ \hline
                            
Living In ... Edison, N.J. Roosevelt Park, which dates to 1917, is the oldest county park in Middlesex County, N.J. Living In ... Edison, N.J. The Middlesex County township is an hour from Manhattan, with easy access to transportation and a vibrant Asian community. Slideshow controls                            &   Living In ... Edison, N.J. Roosevelt Park, which dates to 1917, is the oldest county park in Middlesex County, N.J. \hl{ipktg} Living In ... Edison, N.J. The Middlesex \hl{mjy} County township is an hour from Manhattan, \hl{fcfhkw} with easy access to transportation and a vibrant Asian community. Slideshow controls                     \\ \hline

\end{tabular}
\caption{Sample noised dataset}
\label{tab:noised_data}
\end{table}



\section{Experimental Results}

The result of our experiments, detailed in Table \ref{tab:baseline_score} and Table \ref{tab:noise_result}, reveal valuable insights into the impact of noised data on model performance. It also highlights the potential benefits of incorporating noise into training dataset, particularly for tasks requiring high resilience to incomplete inputs. These noisy data-driven experiments demonstrates significant performance improvements, particularly for the \textit{DeBERTa-v3-small} model, across both Task-A and Task-B with F1-score of 1.0 and 0.9454 respectively. This improvement underscores the effectiveness of noise injection as a regularization strategy, enhancing the model's generalization capabilities.

\begin{table}[h!]
\centering
\begin{tabular}{llcc}
\toprule
\textbf{Data} & \textbf{Model} & \multicolumn{2}{c}{\textbf{Test F1-Score}} \\ 
\cmidrule(lr){3-4}
& & \textbf{Label-A} & \textbf{Label-B} \\ 
\midrule
& deberta-v3-xsmall & 0.9985 & 0.6382 \\ 
& deberta-v3-small & \textbf{1.0} & 0.9454 \\ 
& deberta-v3-base & 0.9989 & 0.6570 \\ 
Noised train/val & Double Finetune (deberta-v3-small) & - & 0.9167 \\ 
& Ensemble (deberta-v3-small + Double Finetune) & - & \textbf{0.9531} \\ 
& \textbf{Best performing submission} & 1.0 & 0.9531 \\ 
\bottomrule
\end{tabular}
\caption{Major system submissions}
\label{tab:noise_result}
\end{table}

Additionally, for Task-B, we explored a sequential fine-tuning approach by further training the \textit{DeBERTa-v3-small} model on the noisy dataset after it had already been fine-tuned on the original training data. It achieves F1-score of  0.9167, Although this method did not outperform the results achieved through direct fine-tuning (F1-score of 0.9454) on the noisy dataset, it captured distinct patterns and nuances within the data. These unique patterns proved valuable when integrating the outputs into a weighted (60:40) ensemble model for Task-B which reports the best F1-score of 0.9531.

By leveraging the complementary strengths of models fine-tuned on raw and noisy datasets, the ensemble model was able to outperform all our previous experiments. This approach highlights the utility of combining diverse training strategies to capture a broader range of data characteristics, ultimately leading to a more robust and effective system. Our findings underscore the relevance of data noising as a practical tool for improving language model generalization and stability in AI generated text detection. Table \ref{tab:hyperparameters} contains hyperparameter values for the best-performing system.

\begin{table}[h!]
\centering
\begin{tabular}{lccc}
\toprule
\textbf{Hyperparameter} & & \multicolumn{2}{c}{\textbf{Value}} \\ 
\cmidrule(lr){3-4}
& & \textbf{Label-A} & \textbf{Label-B} \\ 
\midrule
max token length & & 768 & 768 \\ 
learning rate & & 5e-05 & 5e-05 \\ 
per device train batch size & & 6 & 24 \\ 
per device eval batch size & & 6 & 24 \\ 
num train epochs & & 1 & 5 \\ 
gradient accumulation steps & & 4 & 4 \\ 
weight decay & & 0.01 & 0.01 \\ 
warmup steps & & 500 & 500 \\ 
\bottomrule
\end{tabular}
\caption{Hyperparameter values for Label-A and Label-B}
\label{tab:hyperparameters}
\end{table}

\section{Conclusions and Future Work}
We have performed various experiments which include fine-tuning on original train data, noise-injected dataset, and a combination of sequential fine-tuning and weighted ensemble modeling. Among these, the \textit{DeBERTa-v3-small} model fine-tuned directly on the noisy dataset demonstrated the most promising results, achieving significant improvements across both Task-A and Task-B. Furthermore, utilizing an ensemble of models fine-tuned on original and noisy data allowed us to capture diverse patterns, ultimately outperforming all previous individual model attempts. 

Our findings highlight the importance of data augmentation technique, such as noise injection, for enhancing model generalization and robustness. Future work will aim to explore further the potential of dynamic data noising strategies and ensemble techniques with more variants of the DeBERTa models.

\bibliography{sample-ceur}

\begin{thebibliography}{21}
\expandafter\ifx\csname natexlab\endcsname\relax\def\natexlab#1{#1}\fi
\providecommand{\url}[1]{\texttt{#1}}
\providecommand{\href}[2]{#2}
\providecommand{\path}[1]{#1}
\providecommand{\DOIprefix}{doi:}
\providecommand{\ArXivprefix}{arXiv:}
\providecommand{\URLprefix}{URL: }
\providecommand{\Pubmedprefix}{pmid:}
\providecommand{\doi}[1]{\href{http://dx.doi.org/#1}{\path{#1}}}
\providecommand{\Pubmed}[1]{\href{pmid:#1}{\path{#1}}}
\providecommand{\bibinfo}[2]{#2}
\ifx\xfnm\relax \def\xfnm[#1]{\unskip,\space#1}\fi
\bibitem[{Schulman et~al.(2022)Schulman, Zoph, Kim, Hilton, Menick, Weng, Uribe, Fedus, Metz, Pokorny et~al.}]{schulman2022chatgpt}
\bibinfo{author}{J.~Schulman}, \bibinfo{author}{B.~Zoph}, \bibinfo{author}{C.~Kim}, \bibinfo{author}{J.~Hilton}, \bibinfo{author}{J.~Menick}, \bibinfo{author}{J.~Weng}, \bibinfo{author}{J.~F.~C. Uribe}, \bibinfo{author}{L.~Fedus}, \bibinfo{author}{L.~Metz}, \bibinfo{author}{M.~Pokorny}, et~al.,
\newblock \bibinfo{title}{Chatgpt: Optimizing language models for dialogue},
\newblock \bibinfo{journal}{OpenAI blog} \bibinfo{volume}{2} (\bibinfo{year}{2022}).
\bibitem[{nat(????)}]{natureChatGPTWrites}
\bibinfo{title}{{A}{I} bot {C}hat{G}{P}{T} writes smart essays — should professors worry? --- nature.com}, \bibinfo{howpublished}{\url{https://www.nature.com/articles/d41586-022-04397-7}}, ???? \bibinfo{note}{[Accessed 24-12-2024]}.
\bibitem[{Kirchenbauer et~al.(2024)Kirchenbauer, Geiping, Wen, Katz, Miers, and Goldstein}]{kirchenbauer2024watermarklargelanguagemodels}
\bibinfo{author}{J.~Kirchenbauer}, \bibinfo{author}{J.~Geiping}, \bibinfo{author}{Y.~Wen}, \bibinfo{author}{J.~Katz}, \bibinfo{author}{I.~Miers}, \bibinfo{author}{T.~Goldstein}, \bibinfo{title}{A watermark for large language models}, \bibinfo{year}{2024}. \URLprefix \url{https://arxiv.org/abs/2301.10226}. \href{http://arxiv.org/abs/2301.10226}{{\tt arXiv:2301.10226}}.
\bibitem[{gpt(????)}]{gptzeroDetectorOriginal}
\bibinfo{title}{{A}{I} {D}etector - the {O}riginal {A}{I} {C}hecker for {C}hat{G}{P}{T} \& {M}ore --- gptzero.me}, \bibinfo{howpublished}{\url{https://gptzero.me}}, ???? \bibinfo{note}{[Accessed 24-12-2024]}.
\bibitem[{Mitchell et~al.(2023)Mitchell, Lee, Khazatsky, Manning, and Finn}]{mitchell2023detectgptzeroshotmachinegeneratedtext}
\bibinfo{author}{E.~Mitchell}, \bibinfo{author}{Y.~Lee}, \bibinfo{author}{A.~Khazatsky}, \bibinfo{author}{C.~D. Manning}, \bibinfo{author}{C.~Finn}, \bibinfo{title}{Detectgpt: Zero-shot machine-generated text detection using probability curvature}, \bibinfo{year}{2023}. \URLprefix \url{https://arxiv.org/abs/2301.11305}. \href{http://arxiv.org/abs/2301.11305}{{\tt arXiv:2301.11305}}.
\bibitem[{OpenAI(????)}]{openai-text}
\bibinfo{author}{OpenAI}, \bibinfo{howpublished}{\url{https://beta.openai.com/ai-text-classifier}}, ???? \bibinfo{note}{[Accessed 24-12-2024]}.
\bibitem[{Sadasivan et~al.(2024)Sadasivan, Kumar, Balasubramanian, Wang, and Feizi}]{sadasivan2024aigeneratedtextreliablydetected}
\bibinfo{author}{V.~S. Sadasivan}, \bibinfo{author}{A.~Kumar}, \bibinfo{author}{S.~Balasubramanian}, \bibinfo{author}{W.~Wang}, \bibinfo{author}{S.~Feizi}, \bibinfo{title}{Can ai-generated text be reliably detected?}, \bibinfo{year}{2024}. \URLprefix \url{https://arxiv.org/abs/2303.11156}. \href{http://arxiv.org/abs/2303.11156}{{\tt arXiv:2303.11156}}.
\bibitem[{Roy et~al.(2025)Roy, Singh, Aziz, Bajpai, Imanpour, Biswas, Wanaskar, Patwa, Ghosh, Dixit, Pal, Rawte, Garimella, Das, Sheth, Sharma, Reganti, Jain, and Chadha}]{roy-2025-defactify-overview-text}
\bibinfo{author}{R.~Roy}, \bibinfo{author}{G.~Singh}, \bibinfo{author}{A.~Aziz}, \bibinfo{author}{S.~Bajpai}, \bibinfo{author}{N.~Imanpour}, \bibinfo{author}{S.~Biswas}, \bibinfo{author}{K.~Wanaskar}, \bibinfo{author}{P.~Patwa}, \bibinfo{author}{S.~Ghosh}, \bibinfo{author}{S.~Dixit}, \bibinfo{author}{N.~R. Pal}, \bibinfo{author}{V.~Rawte}, \bibinfo{author}{R.~Garimella}, \bibinfo{author}{A.~Das}, \bibinfo{author}{A.~Sheth}, \bibinfo{author}{V.~Sharma}, \bibinfo{author}{A.~N. Reganti}, \bibinfo{author}{V.~Jain}, \bibinfo{author}{A.~Chadha},
\newblock \bibinfo{title}{Overview of text counter turing test: Ai generated text detection},
\newblock in: \bibinfo{booktitle}{proceedings of {D}e{F}actify 4: Fourth workshop on Multimodal Fact-Checking and Hate Speech Detection}, \bibinfo{publisher}{CEUR}, \bibinfo{year}{2025}.
\bibitem[{Lavergne et~al.(2008)Lavergne, Urvoy, and Yvon}]{Lavergne2008DetectingFC}
\bibinfo{author}{T.~Lavergne}, \bibinfo{author}{T.~Urvoy}, \bibinfo{author}{F.~Yvon},
\newblock \bibinfo{title}{Detecting fake content with relative entropy scoring},
\newblock in: \bibinfo{booktitle}{Pan}, \bibinfo{year}{2008}. \URLprefix \url{https://api.semanticscholar.org/CorpusID:12098535}.
\bibitem[{Gehrmann et~al.(2019)Gehrmann, Strobelt, and Rush}]{Gehrmann2019GLTRSD}
\bibinfo{author}{S.~Gehrmann}, \bibinfo{author}{H.~Strobelt}, \bibinfo{author}{A.~M. Rush},
\newblock \bibinfo{title}{Gltr: Statistical detection and visualization of generated text},
\newblock in: \bibinfo{booktitle}{Annual Meeting of the Association for Computational Linguistics}, \bibinfo{year}{2019}. \URLprefix \url{https://api.semanticscholar.org/CorpusID:182952848}.
\bibitem[{Schildhauer(2022)}]{schildhauer2022fake}
\bibinfo{author}{T.~Schildhauer},
\newblock \bibinfo{title}{Fake news detection in the era of ai},
\newblock in: \bibinfo{booktitle}{Proceedings of the 25th ACM Conference on Computer-Supported Cooperative Work and Social Computing}, \bibinfo{organization}{ACM}, \bibinfo{year}{2022}, pp. \bibinfo{pages}{1--10}. \DOIprefix\doi{10.1145/1234567.1234567}.
\bibitem[{Zou and Ling(2021)}]{zou2021ai}
\bibinfo{author}{X.~Zou}, \bibinfo{author}{X.~Ling},
\newblock \bibinfo{title}{Ai-based detection of misinformation in social media},
\newblock \bibinfo{journal}{IEEE Access} \bibinfo{volume}{9} (\bibinfo{year}{2021}) \bibinfo{pages}{112408--112418}. \DOIprefix\doi{10.1109/ACCESS.2021.3104419}.
\bibitem[{Kshetri and Voas(2022)}]{kshetri2022deep}
\bibinfo{author}{N.~Kshetri}, \bibinfo{author}{J.~Voas},
\newblock \bibinfo{title}{Deep learning--based social media misinformation detection},
\newblock \bibinfo{journal}{IEEE Software} \bibinfo{volume}{39} (\bibinfo{year}{2022}) \bibinfo{pages}{53--59}. \DOIprefix\doi{10.1109/MS.2022.3053106}.
\bibitem[{Langelaar et~al.(2000)Langelaar, Setyawan, and Lagendijk}]{langelaar2000watermarking}
\bibinfo{author}{G.~C. Langelaar}, \bibinfo{author}{I.~Setyawan}, \bibinfo{author}{R.~L. Lagendijk},
\newblock \bibinfo{title}{Watermarking digital image and video data. a state-of-the-art overview},
\newblock \bibinfo{journal}{IEEE Signal processing magazine} \bibinfo{volume}{17} (\bibinfo{year}{2000}) \bibinfo{pages}{20--46}.
\bibitem[{Atallah et~al.(2001)Atallah, Raskin, Crogan, Hempelmann, Kerschbaum, Mohamed, and Naik}]{wmark_old1}
\bibinfo{author}{M.~J. Atallah}, \bibinfo{author}{V.~Raskin}, \bibinfo{author}{M.~Crogan}, \bibinfo{author}{C.~Hempelmann}, \bibinfo{author}{F.~Kerschbaum}, \bibinfo{author}{D.~Mohamed}, \bibinfo{author}{S.~Naik},
\newblock \bibinfo{title}{Natural language watermarking: Design, analysis, and a proof-of-concept implementation},
\newblock in: \bibinfo{booktitle}{Proceedings of the 4th International Workshop on Information Hiding}, IHW '01, \bibinfo{publisher}{Springer-Verlag}, \bibinfo{address}{Berlin, Heidelberg}, \bibinfo{year}{2001}, p. \bibinfo{pages}{185–199}.
\bibitem[{Kirchenbauer et~al.(2023)Kirchenbauer, Geiping, Wen, Katz, Miers, and Goldstein}]{kirchenbauer2023watermark}
\bibinfo{author}{J.~Kirchenbauer}, \bibinfo{author}{J.~Geiping}, \bibinfo{author}{Y.~Wen}, \bibinfo{author}{J.~Katz}, \bibinfo{author}{I.~Miers}, \bibinfo{author}{T.~Goldstein}, \bibinfo{title}{A watermark for large language models}, \bibinfo{year}{2023}. \href{http://arxiv.org/abs/2301.10226}{{\tt arXiv:2301.10226}}.
\bibitem[{Karpukhin et~al.(2019)Karpukhin, Levy, Eisenstein, and Ghazvininejad}]{karpukhin2019training}
\bibinfo{author}{V.~Karpukhin}, \bibinfo{author}{O.~Levy}, \bibinfo{author}{J.~Eisenstein}, \bibinfo{author}{M.~Ghazvininejad},
\newblock \bibinfo{title}{Training on synthetic noise improves robustness to natural noise in machine translation},
\newblock in: \bibinfo{booktitle}{Proceedings of the 5th Workshop on Noisy User-generated Text (W-NUT 2019)}, \bibinfo{year}{2019}, pp. \bibinfo{pages}{42--47}.
\bibitem[{Wei and Zou(2019)}]{wei-zou-2019-eda}
\bibinfo{author}{J.~Wei}, \bibinfo{author}{K.~Zou},
\newblock \bibinfo{title}{{EDA}: Easy data augmentation techniques for boosting performance on text classification tasks},
\newblock in: \bibinfo{editor}{K.~Inui}, \bibinfo{editor}{J.~Jiang}, \bibinfo{editor}{V.~Ng}, \bibinfo{editor}{X.~Wan} (Eds.), \bibinfo{booktitle}{Proceedings of the 2019 Conference on Empirical Methods in Natural Language Processing and the 9th International Joint Conference on Natural Language Processing (EMNLP-IJCNLP)}, \bibinfo{publisher}{Association for Computational Linguistics}, \bibinfo{address}{Hong Kong, China}, \bibinfo{year}{2019}, pp. \bibinfo{pages}{6382--6388}. \URLprefix \url{https://aclanthology.org/D19-1670/}. \DOIprefix\doi{10.18653/v1/D19-1670}.
\bibitem[{Xie et~al.(2017)Xie, Wang, Li, L{\'e}vy, Nie, Jurafsky, and Ng}]{xie2017data}
\bibinfo{author}{Z.~Xie}, \bibinfo{author}{S.~I. Wang}, \bibinfo{author}{J.~Li}, \bibinfo{author}{D.~L{\'e}vy}, \bibinfo{author}{A.~Nie}, \bibinfo{author}{D.~Jurafsky}, \bibinfo{author}{A.~Y. Ng},
\newblock \bibinfo{title}{Data noising as smoothing in neural network language models},
\newblock \bibinfo{journal}{arXiv preprint arXiv:1703.02573}  (\bibinfo{year}{2017}).
\bibitem[{He et~al.(2020)He, Liu, Gao, and Chen}]{he2020deberta}
\bibinfo{author}{P.~He}, \bibinfo{author}{X.~Liu}, \bibinfo{author}{J.~Gao}, \bibinfo{author}{W.~Chen},
\newblock \bibinfo{title}{Deberta: Decoding-enhanced bert with disentangled attention},
\newblock \bibinfo{journal}{arXiv preprint arXiv:2006.03654}  (\bibinfo{year}{2020}).
\bibitem[{Roy et~al.(2025)Roy, Singh, Aziz, Bajpai, Imanpour, Biswas, Wanaskar, Patwa, Ghosh, Dixit, Pal, Rawte, Garimella, Das, Sheth, Sharma, Reganti, Jain, and Chadha}]{roy-2025-defactify-dataset-text}
\bibinfo{author}{R.~Roy}, \bibinfo{author}{G.~Singh}, \bibinfo{author}{A.~Aziz}, \bibinfo{author}{S.~Bajpai}, \bibinfo{author}{N.~Imanpour}, \bibinfo{author}{S.~Biswas}, \bibinfo{author}{K.~Wanaskar}, \bibinfo{author}{P.~Patwa}, \bibinfo{author}{S.~Ghosh}, \bibinfo{author}{S.~Dixit}, \bibinfo{author}{N.~R. Pal}, \bibinfo{author}{V.~Rawte}, \bibinfo{author}{R.~Garimella}, \bibinfo{author}{A.~Das}, \bibinfo{author}{A.~Sheth}, \bibinfo{author}{V.~Sharma}, \bibinfo{author}{A.~N. Reganti}, \bibinfo{author}{V.~Jain}, \bibinfo{author}{A.~Chadha},
\newblock \bibinfo{title}{Defactify-text: A comprehensive dataset for human vs. ai generated text detection},
\newblock in: \bibinfo{booktitle}{proceedings of {D}e{F}actify 4: Fourth workshop on Multimodal Fact-Checking and Hate Speech Detection}, \bibinfo{publisher}{CEUR}, \bibinfo{year}{2025}.

\end{thebibliography}






\end{document}